\pdfoutput=1

\documentclass[11pt]{article}

\usepackage[final]{acl}
\usepackage{circledsteps}

\usepackage{times}
\usepackage{latexsym}

\usepackage[T1]{fontenc}

\usepackage[utf8]{inputenc}

\usepackage{microtype}

\usepackage{inconsolata}
\usepackage{graphicx}
\usepackage{enumitem}
\usepackage{hyperref}   
\usepackage[nameinlink,capitalise]{cleveref}
\usepackage{multirow}
\usepackage{xcolor}         
\usepackage{makecell}
\usepackage{graphicx}
\usepackage{csquotes}
\usepackage{caption} 
\usepackage{accents}
\usepackage{wrapfig}
\usepackage{soul}
\usepackage{subcaption}
\usepackage{listings}
\usepackage{hyperref}       
\usepackage{url}            
\usepackage{booktabs}
\usepackage{amsfonts}
\usepackage{lipsum}         
\usepackage{multirow}
\usepackage{graphicx}

\title{Multimodal Self-Instruct: Synthetic Abstract Image and Visual Reasoning Instruction Using Language Model}

\author{Wenqi Zhang$^{1, *}$, Zhenglin Cheng$^{1, *}$, Yuanyu He$^{1}$, Mengna Wang$^{2}$, Yongliang Shen$^{1}$\\
       {\bf Zeqi Tan$^{1}$, Guiyang Hou$^{1}$, Mingqian He$^{1}$, Yanna Ma$^{3}$, Weiming Lu$^{1, \dagger}$, Yueting Zhuang$^{1}$}\\
  $^1$College of Computer Science and Technology, Zhejiang University\\
  $^2$Institute of Software, Chinese Academy of Sciences\\ 
  $^3$University of Shanghai for Science and Technology\\
  \texttt{\{zhangwenqi, luwm\}@zju.edu.cn}\\
  Project Page: \url{https://multi-modal-self-instruct.github.io}}

\begin{document}
\maketitle
\renewcommand{\thefootnote}{\fnsymbol{footnote}} 
\footnotetext[1]{The first two authors have equal contributions.} 
\footnotetext[2]{Corresponding author.}  

\renewcommand{\thefootnote}{\arabic{footnote}}

\begin{abstract}
Although most current large multimodal models (LMMs) can already understand photos of natural scenes and portraits, their understanding of abstract images, e.g., charts, maps, or layouts, and visual reasoning capabilities remains quite rudimentary. They often struggle with simple daily tasks, such as reading time from a clock, understanding a flowchart, or planning a route using a road map. In light of this, we design a multi-modal self-instruct pipeline, utilizing large language models and their code capabilities to synthesize massive abstract images and visual reasoning instructions across daily scenarios. Our strategy effortlessly creates a multimodal benchmark with 11,193 instructions for eight visual scenarios: charts, tables, simulated maps, dashboards, flowcharts, relation graphs, floor plans, and visual puzzles. \textbf{This benchmark, constructed with simple lines and geometric elements, exposes the shortcomings of most advanced LMMs} like Claude-3.5-Sonnet and GPT-4o in abstract image understanding, spatial relations reasoning, and visual element induction. Besides, to verify the quality of our synthetic data, we fine-tune an LMM using 62,476 synthetic chart, table and road map instructions.
The results demonstrate improved chart understanding and map navigation performance, and also demonstrate potential benefits for other visual reasoning tasks. Our code is available at: \url{https://github.com/zwq2018/Multi-modal-Self-instruct}.

\end{abstract}

\begin{figure}[!htb] 
\centering 
\includegraphics[width=0.45\textwidth]{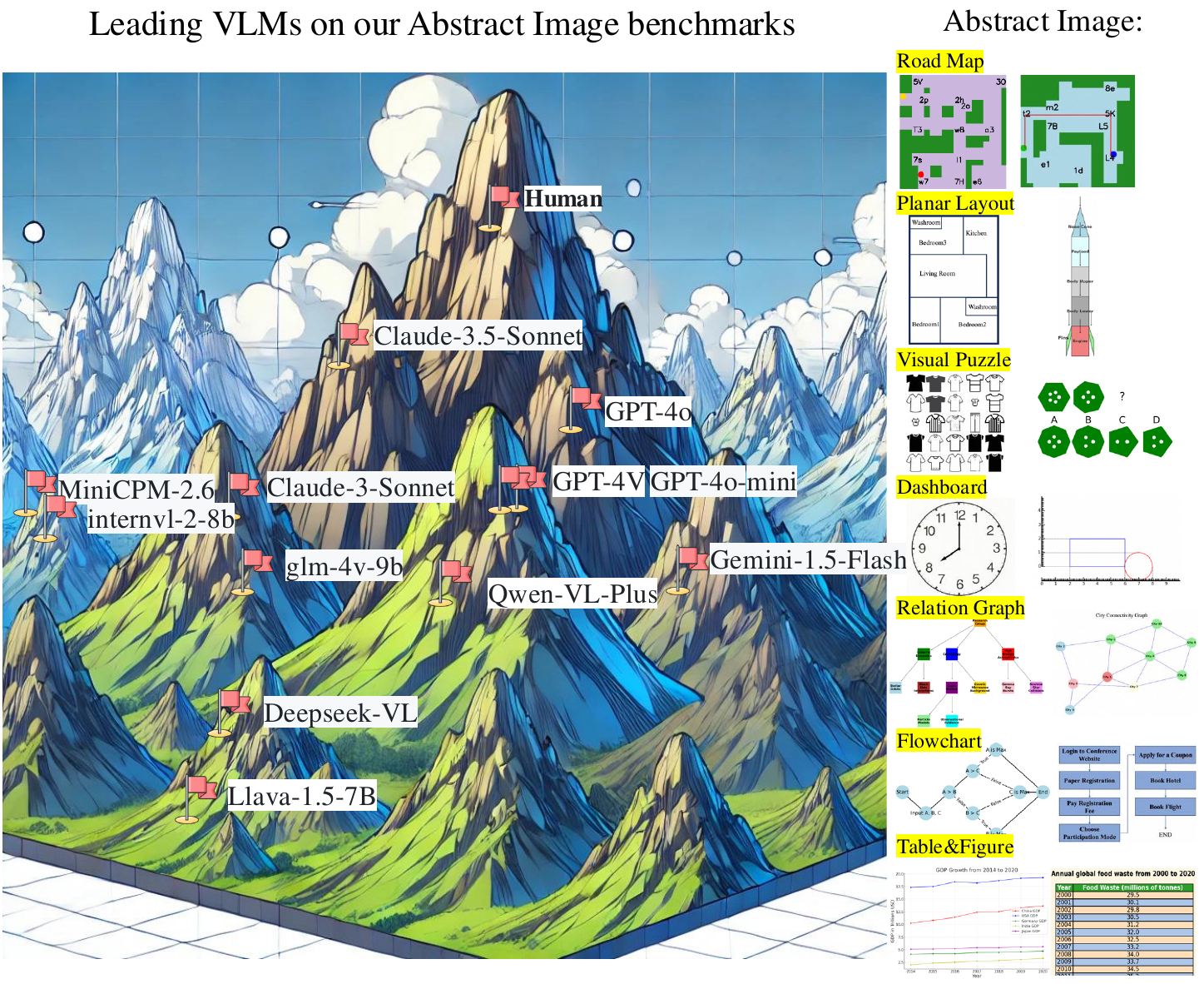} %
\caption{Benchmarking Leading LMMs on abstract image understanding and reasoning tasks.}\label{figure0} 
\end{figure}

\begin{figure*}[!htb] 
\centering 
\includegraphics[width=0.9\textwidth]{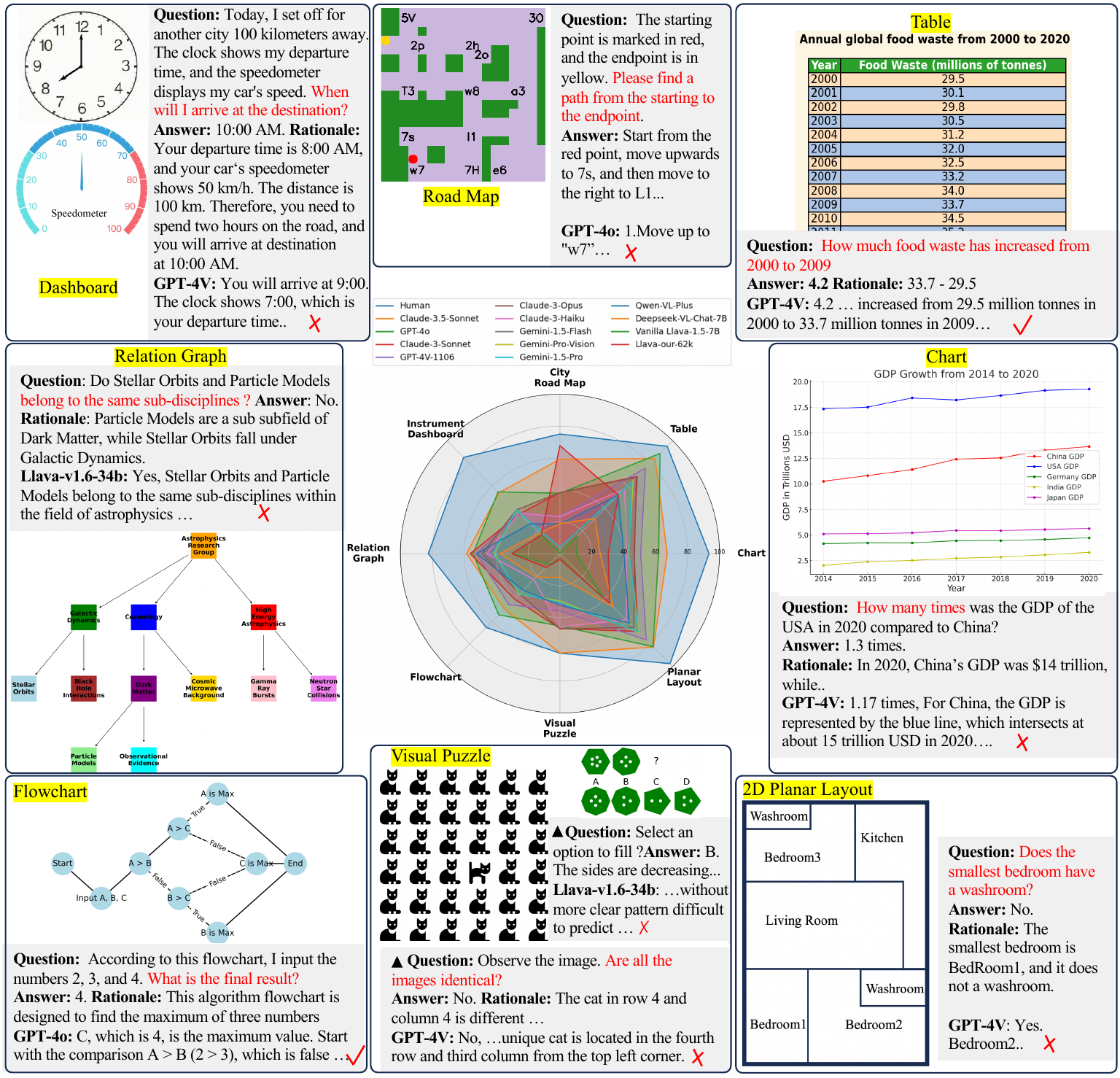} %
\caption{We leverage LLM and code to synthesize abstract images and self-instruct diverse reasoning instructions, e.g., charts, road maps, dashboards, visual puzzles, and relation graphs. Unlike natural landscapes and human photos, these non-natural images constructed with geometric elements require stronger perception and spatial relation reasoning. Our benchmark indicates that current LMMs are far from human-level performance. They even fail to complete simple daily tasks, e.g., reading the time on a clock or planning a route using a map.}\label{figure1} 
\end{figure*}


\section{Introduction}
In recent times, spurred by breakthroughs in large language models (LLMs)~\citep{Zeng2023GLM, touvron2023llama, Chatgpt, gpt4, Touvron2023Llama2O,bi2024deepseek,jiang2024mixtral,anthropic2024claude,abdin2024phi}, large multimodal models (LMMs) have also undergone rapid advancements~\citep{liu2024visual, liu2023improved, team2023gemini,bai2023qwen,lu2024deepseek,mckinzie2024mm1}. Leveraging a pre-trained LLM to encode all modalities empowers LMMs to understand human daily environments and execute complex tasks~\citep{hong2023cogagent,Zhang2023DataCopilotBB,hu2023mplug,yang2023appagent,zhang-etal-2024-self-contrast, koh2024visualwebarena,zhang-etal-2024-agent}. This greatly expands the potential of general-purpose AI assistants.

Despite these achievements, LMMs still exhibit significant deficiencies when deployed in human daily life~\citep{yin2023survey,xie2024large}. For instance, LMMs often fail when planning a route using a road map, reading the time from a clock image, or interpreting a flowchart. We observe that these simple daily activities require LMMs to understand abstract images, such as maps, charts, and dashboards, rather than natural photographs or portraits with explicit semantics. These abstract images composed of simple geometric elements are more challenging for LMMs. Furthermore, even many advanced LMMs are easily stumped by simple visual-level reasoning tasks, such as geometric pattern induction and visual symbol comparison.






However, these capabilities, i.e., perceiving abstract images and reasoning about visual elements, are essential for LMMs if we deploy an LMM-driven agent in our daily lives. It can help us with data analysis, map navigation, web searches, and many other tedious tasks. On the one hand, despite valuable explorations by some pioneers~\citep{yu2023mm,liu2023mmbench, han2023chartllama,ying2024mmt,wei2024mchartqa}, these abstract image understanding and visual reasoning abilities have not been adequately emphasized, and we need a dedicated benchmark to systematically evaluate the performance of current LMMs in this aspect. On the other hand, unlike semantic-related tasks, collecting such abstract image-text pairs with reasoning context is labor-intensive and time-consuming.









To fill in the gap, we drew inspiration from synthetic data~\citep{wang2022selfinstruct, liu2024best,han2023chartllama,du2023makes}, which is widely used to supplement the insufficiency of instruction-following data. For instance, distilling high-quality dialogue data from a strong LLM~\citep{wang2022selfinstruct,xu2023wizardlm,yu2023metamath,chen2023sharegpt4v,zhao2023genixer}, or using external tools to refine the quality of synthetic data~\citep{wei2023magicoder,lee2024llm2llm}. However, synthesizing image-text data for LMM is not easy, as current LLMs can not directly generate images. An intuitive approach is to combine LLMs with a text-to-image model for producing <image, question, answer>~\citep{li2023stablellava,wu2023datasetdm}, but most text-to-image models fail to finely control the details of the image~\citep{betker2023improving,esser2024scaling}, potentially leading to a misalignment between image and text. 



Considering that abstract images are composed of lines and geometric elements, we can utilize code to accurately synthesize them. In light of this, we advocate a code-centric self-instruct strategy to synthesize massive abstract images with reasoning questions and answer pairs. We first instruct LLM to autonomously propose a creative visual idea for a daily scenario and then self-propose the necessary data and code to draw an abstract image, such as plotting a relation graph or house layout. After synthesizing images, our strategy self-instructs multiple reasoning question-answer pairs based on the plotting idea and code. This code-centric design can effortlessly synthesize diverse abstract images and reasoning instructions, involving chart interpretation, spatial relation reasoning, visual puzzles, and mathematical geometry problems, and also provide accurate answers and rationale.


As shown in~\Cref{figure1}, our strategy synthesized an abstract image benchmark for daily scenarios, including 11,193 high-quality instructions covering eight scenarios: Dashboard, Road Map, Chart, Table, Flowchart, Relation Graph, Visual Puzzles, and 2D Planar Layout. Empowered by this benchmark, we evaluate several representative LMMs and identify their significant deficiencies in abstract image understanding and visual reasoning. For example, in the dashboard scene, the best-performing LMM (GPT-4o) only achieved a score of 54.7, far below the human level of 85.3. Our abstract image benchmark further indicates that the gap between current open-source models and closed-source models remains significant, despite their comparable performance on semantics-related benchmarks. 

Besides, to verify the quality of the synthesized data, we synthesized 62,476 charts and road map instructions for fine-tuning Llava-1.5-7B. Experimental results show that our synthesized data can significantly enhance in-domain performance and also benefit other abstract image reasoning tasks.

Our contributions can be summarized as follows:
\begin{itemize}
    \item We identify that current LMMs have a significant gap compared to humans in understanding and visually reasoning about abstract images, such as maps, charts, and layouts. 
    
    \item Utilizing LLM and code, We design a multi-modal self-instruct strategy to synthesize a diverse set of abstract images and reasoning instructions, providing value data for LMMs.
    
    
    \item We synthesized a benchmark of 11,193 high-quality abstract images, covering eight common scenarios. Our benchmark reveals significant deficiencies even in advanced LMMs. Besides, we synthesized 62,476 chart and road map instructions for fine-tuning, verifying the effectiveness of the synthesized data.


\end{itemize}


\begin{figure*}[!htb] 
\centering 
\includegraphics[width=0.9\textwidth]{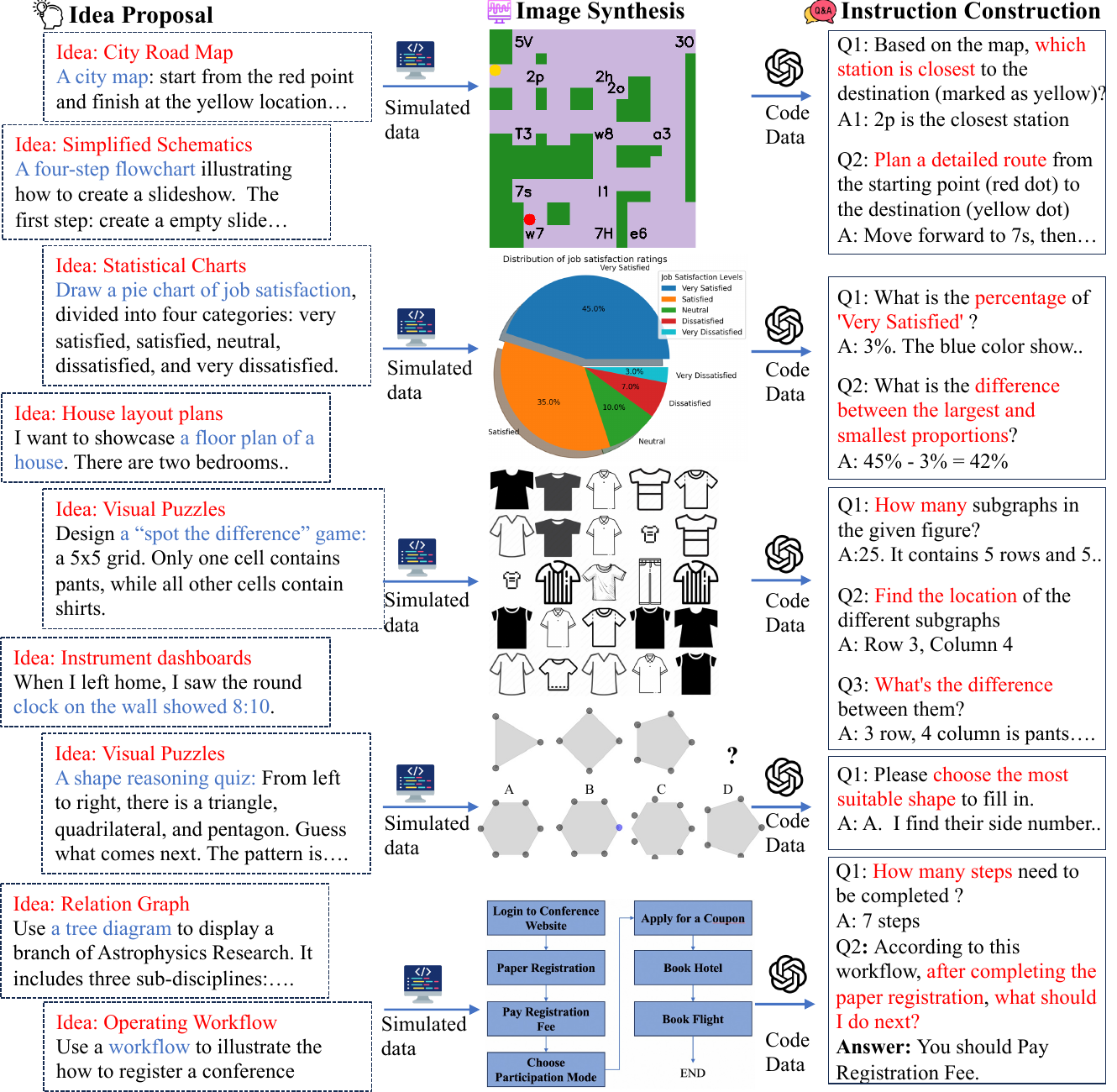} %
\caption{Our multi-modal self-instruct strategy first self-proposes a visual idea to depict an abstract image. Based on this, the LLM generates simulated data and writes code to create the drawings. Subsequently, LLM is instructed to design multiple Q\&A based on the code and idea, covering various aspects such as spatial reasoning, color recognition, and mathematical reasoning, constructing a rich set of multimodal instructions.}\label{figure2} 
\end{figure*}

\section{Multi-modal Self-Instruct}
\subsection{Overview}
Our multi-modal self-instruct is an LLM-driven data synthesis strategy capable of producing abstract images and aligned reasoning instructions for various daily scenarios, including road maps, dashboards, 2D planar layouts, charts, relation graphs, flowcharts, and visual puzzles.



Firstly, our strategy can autonomously propose a creative idea for visual scenarios, e.g., \textit{using a step-by-step flowchart to demonstrate how to attend an academy conference} or \textit{designing road map} (\Cref{Visual Idea Proposal}). Then it generates detailed code to visualize this idea (\Cref{Visual Content Creation}). After synthesizing the desired image, LLMs self-instruct multiple high-quality Q\&A pairs for this visual content (\Cref{Visual Instruction Construction}). The entire process is fully completed by the LLM with a few demonstrations.

As shown in~\Cref{figure2}, we illustrate the entire process of our image-text synthesis, including using road maps for navigation, interpreting pie charts, solving visual puzzles, and using operating workflow. For each scenario, we synthesize multiple questions, annotated answers, and rationales. For example, in the pie chart case, the LLM designs a multi-step math question about the difference between the largest and smallest categories. 


\subsection{Visual Idea Proposal} \label{Visual Idea Proposal}

To generate an image from scratch, we first instruct the LLM to propose an innovative visual idea. This visual idea illustrates a scenario commonly encountered in daily life or work, e.g., a chart about a specific topic or a road map. Besides, this scenario image can be rendered with code, rather than real portraits or natural scenes. Therefore, we focus on eight common types of abstract images that are rarely covered in current datasets: 
\lstset{
    framesep = 20pt,
    rulesep = 10pt,
    backgroundcolor = \color[RGB]{245,245,244},
    breaklines = true,
    breakindent = 0pt,
    basicstyle = \ttfamily\small,
    escapeinside = {(*@}{@*)} 
}
\begin{lstlisting}
(*@\color{blue}{Working Scene and Life Scene}@*)
(*@\textbf{Charts and Table}@*): Line, bar, pie, composite charts, and single and multiple tables. 
(*@\textbf{Flowchart}@*): Algorithm flowcharts and operating workflows, such as designing a slide presentation.
(*@\textbf{Relation Graph}@*): Multiple relational graphs with complex connections.
(*@\textbf{Road Map}@*): Simulated road maps annotated with intersection names.
(*@\textbf{Visual Puzzles}@*): 1. Inductive reasoning across multiple images. 2. Comparing the differences between multiple images.
(*@\textbf{2D Planar Layout}@*): Floor plans with different structures and layouts.
(*@\textbf{Instrument Dashboards}@*): Mechanical dials, such as clocks, odometers, speedometers, thermometers, barometers..
\end{lstlisting} 

We design some examples for each scenario as in-context demonstrations. Prompted by them, the LLM is encouraged to propose a creative and detailed plotting idea using natural language.
These visual ideas depict the basic outlines of visual information. By incorporating detailed parameters, a visual idea can control the specifics of image synthesis, enabling the creation of a diverse range of images.
Additionally, when constructing visual instructions, visual ideas can provide a visual reference for the generation of instructions in natural language form.

\subsection{Image Synthesis} \label{Visual Content Creation}
\paragraph{Simulated Data} To render the proposed idea into an image, we guide the LLM to first generate some simulated data for the proposed idea. For example, for the pie chart in \Cref{figure2}, the LLM needs to fabricate the percentage data for the four types.

\paragraph{Code Generation}
After producing simulated data, LLM generates corresponding Python code to visualize the proposed idea. We encourage the LLM to use popular visualization packages, e.g., Matplotlib\footnote{https://matplotlib.org} or ECharts\footnote{https://echarts.apache.org/zh/index.html}, to create desired visual elements, as it significantly reduces the complexity of code generation. Besides, we instruct the LLM to explicitly define all parameters in the code for plotting images, such as image style, color, font size, and legend position. These explicitly stated parameters control the details of the synthesized images and can be used to produce Q\&A.

\subsection{Visual Instruction Construction} \label{Visual Instruction Construction}
After executing the code, we obtain the expected image. Next, the LLM autonomously proposes multiple high-quality <question, answer> pairs related to this synthetic image.

\paragraph{Question-Answer Pair Generation.} To make the LLM aware of all the image details, we concatenate the proposed idea, simulated data, and generated code in the prompt, and then guide the LLM to design instructions following data for this synthesized image. More than just image comprehension and captioning tasks, our strategy can self-propose a wide range of unconventional questions for this synthesized image, such as comparing differences among multiple images, area estimation, and spatial relation inference. Furthermore, it can even design diverse multi-step reasoning problems based on multiple synthesized images.

\paragraph{Annotate Answers with Rationale.}
To enhance the training effectiveness of multimodal instruction-following data, we also provide a detailed rationale for each question. We prompt the LLM to carefully review the idea and code, and then generate a detailed rationale for the given question, rather than just providing an answer. Similar to the chain-of-thought process, rationale can be used to train LMMs, enhancing their reasoning capabilities.

Below is a complete case for our pipeline, including Idea Proposal, Image Synthesis, and Instruction Construction. We also provide the results of GPT-4 and Gemini-1.5, which all failed on this case.

\begin{lstlisting}
(*@\color{blue}{Idea Proposal}@*): Draw a clock with hour and minute hands.
(*@\color{blue}{Simulated Data}@*): time='8:10', Shape='Round Clock', color='black', size=...
(*@\color{blue}{Code Generation}@*): 'import pyechart...'
(*@\color{blue}{Instruction Construction}@*)
(*@\textbf{Question}@*): What time is shown on the dial?
(*@\textbf{Answer1: 8:10}@*)
(*@\color{red}{GPT-4V: 10:10}@*). (*@\color{red}{Gemini-1.5-pro: 2:42}@*). 
(*@\textbf{Math Question}@*): When I left home, the clock showed the time indicated in the figure. What time is it after 8 hours of work?
(*@\textbf{Answer2: 4:10 or 16:10}@*)
(*@\textbf{Rationale}@*): I see that the clock shows the time as 8:10. After working for eight hours, the time should be 16:10.
(*@\color{red}{GPT-4V: 7:10}@*). The clock shows 11:10 ... 
(*@\color{red}{Gemini-1.5-pro: 9:50}@*). The time is 1:50 ...
(*@\textbf{Reasoning Question}@*): I exercised for one and a half hours. After finishing, the clock showed the time as illustrated. What number did the hour hand point to when I started my workout?
(*@\textbf{Answer3: 6 or 7}@*)
(*@\textbf{Rationale}@*): I read the time from the clock as 8:10, and you have been exercising for an hour and a half. This means you left at 6:40. Therefore ...
(*@\color{red}{GPT-4V: 12.}@*) The clock shows the time as 1:30 ... 1:30-1.5 hours=12:00 PM ...
(*@\color{red}{Gemini-1.5-pro: 1.}@*) The clock is 2:30 ... An hour and a half before was 1:00 ...
\end{lstlisting}


\begin{table}[t!]\small
    \centering
    \setlength\tabcolsep{2pt} 

    \begin{tabular}{ l c  c c c}
    \toprule[1pt]
    \textbf{\makecell[l]{Task}} & \textbf{\textbf{\#Image}}  & \textbf{\makecell[c]{\# Instruction}} & \textbf{\makecell[c]{\#Usage} } \\ \midrule
    Chart            & 1,768  &  34,590  &     Train   \\ 
    Table            & 570    &  10,886  &     Train  \\ 
    Road map         & 17,000 &  17,000  &     Train   \\ 
    \textbf{All}              & 19,338 &  62,476  &      Train   \\
    \midrule
    Chart            & 149 & 3,018 &     Test   \\ 
    Table            & 58 &  1,108  &     Test  \\ 
    Road map         & 3,000 &  3,000  &     Test  \\ 
    Dashboard        & 73 & 1,013 &     Test  \\
    Relation Graph   & 66 &  822  &     Test  \\ 
    Flowchart        & 98 &  1,451 &     Test   \\ 
    Visual Puzzle    & 189 & 529 &     Test   \\ 
    2D Planar Layout & 25 & 252 &     Test    \\
    \textbf{All}              & 3,658 & 11,193 & Test \\
    \bottomrule[1pt]
    \end{tabular}
    \caption{The statistics of our dataset, including eight tasks from work and life scenarios. All data were synthesized using our multi-modal self-instruct strategy.}\label{table statistics}
\end{table}

\begin{figure}[!t] 
\centering 
\includegraphics[width=0.45\textwidth]{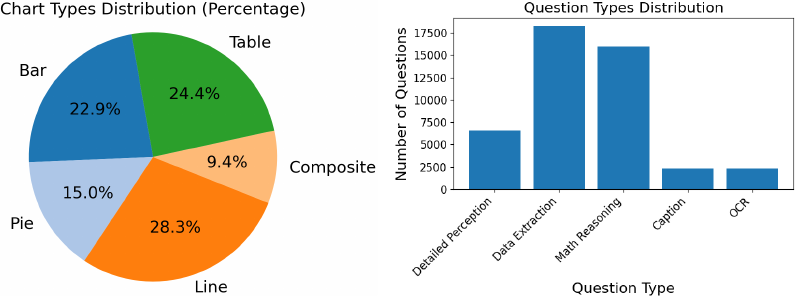} %
\caption{Left: The distribution of different chart types. Right: The number of questions for each category.}\label{chart and question type} 
\end{figure}

\section{Multimodal Self-instruct Dataset}
\subsection{Dataset Statistics}
We focus on eight common but under-explored scenario images, including Chart, Table, Road Map, Relation Graph, Flowchart, Visual Puzzle, Dashboard, and 2D Planar Layout. We initially synthesized a benchmark involving all 8 scenarios, containing 3,658 images and 11,193 instructions in total, to benchmark several representative LMMs. Besides, to evaluate the quality of the synthesized data, we also synthesize three training sets for chart, table, and road map tasks, comprising 34,590, 10,886, and 17,000 training instructions, respectively. As shown in \Cref{table statistics}, we provide detailed statistics about our synthesized dataset.



\begin{figure}[!t] 
\centering 
\includegraphics[width=0.4\textwidth]{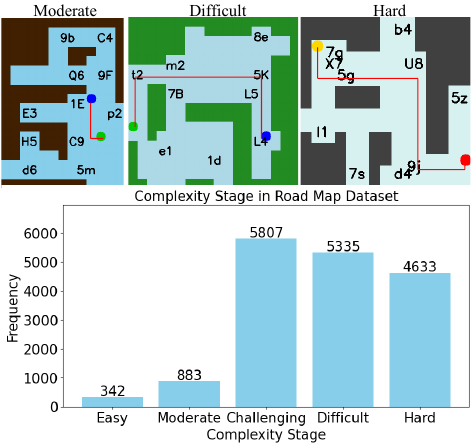} %
\caption{Top: We present three examples of road maps with different path complexity. Bottom: We categorize all maps into five levels of complexity.}\label{figure4} 
\end{figure}

\subsection{Synthesis Details} \label{dataset detail}
\paragraph{Chart and Table} Firstly, we design some keyword seeds, e.g., GDP, energy consumption, employment rate, and then we prompt the LLM to expand these seed keywords into a huge keyword library covering economics, technology, and society domains. Before generation, we first randomly sample a keyword from the library and then prompt the LLM to generate corresponding visual ideas, code, and instruction data. We synthesize five types of charts: \textit{line charts, bar charts, pie charts, table screenshots, and composite charts (containing multiple sub-charts)}. For each chart, we prompt LLMs to self-instruct five types of questions: \textit{Optical Character Recognition (OCR), Caption, Detailed Perception (involving issues of position, quantity, layout), Data Extraction, and Mathematical Reasoning}. As shown in~\Cref{chart and question type}, we provide statistics based on chart types and question types separately. Besides, we provide several detailed examples for each type of chart and question in~\Cref{chart example}. 

\paragraph{Road map Navigation.} To generate simulated maps with obstacles and paths, we design a path generation strategy based on the rapidly exploring random tree algorithm: Starting from an initial point, the agent randomly walks within an under-explored map, sampling the path according to the predefined walking parameters, including direction, probability, and maximum walking steps. The process stops when the maximum walking steps are reached, and the stopping position is set as the endpoint. When synthesizing maps, the LLM first sets the map size, and randomly walking parameters. Then it generates code to implement our path generation process. Ultimately, we synthesized 17$k$ training maps and 3$k$ testing maps. Based on the path complexity, we categorized all maps into five levels. As shown in~\Cref{figure4}, most maps are of medium difficulty or higher, requiring at least two intersections and turns to reach the endpoint. We provide two complete cases in~\Cref{map navigation example}.

\paragraph{Other Scenarios Synthesis.} We employ similar processes to synthesize images of the other five scenarios, producing 1,013 Dashboard, 822 Relation Graph, 1,451 Flowchart, 529 Visual Puzzle, and 252 2D Planar Layout instructions. Specifically, for Flowchart, we synthesize two types: algorithm flowcharts and operating workflow. For the Relation Graph, we generate graphs with different structures, such as trees or graphs. For Dashboard, we synthesize circular dials, such as clocks, speedometers, and fuel gauges, and some elongated dials like thermometers and barometers. Regarding the Visual Puzzle task, we synthesize two types of puzzles: visual pattern induction and multi-subgraph comparison. As for the 2D Planar Layout, we synthesize architectural layouts, webpage layouts, and more. These instructions are all used as test benchmarks to evaluate the current mainstream LMMs performance. We provide some visualized cases for each task in~\Cref{flowchart example,relation graph example,layout example,visual puzzle example}.

\subsection{Implementation Details} \label{Implementation detail}
\textbf{LLM and Prompts.} We employ \textit{gpt-4-turbo-2024-04-09} to implement our data synthesis: idea proposal, code generation, and instruction construction. A detailed prompt is shown in~\Cref{detail prompt}.

\paragraph{Dataset Diversity.} Firstly, in the data synthesis process, we control the generated topic of the image with many pre-defined keywords. For example, before synthesizing the chart, we designed a keyword library (e.g., GDP, energy, and employment rate) that includes various keywords from different domains covering economics, technology, and society. This strategy can control the generated content and avoid deviations. Similarly, during the image and question synthesis process, we use few-shot examples and templates to control the types of questions and images generated. For example, we generate five types of charts (bar, table, line, pie, composite) for the chart task, and also 5 types of questions (perception, extraction, math, caption, OCR). We also generate the difficulty levels of synthesized maps. The quantity for each category can be predefined in advance.

\paragraph{Dataset Quality.} To ensure the quality of the synthesized data, we filtered the data at three levels: \textbf{code feasibility, image aesthetics, and answer accuracy}. I. If the generated code fails to run, we prompt the LLM to self-reflect based on the error feedback from the compiler. If the LLM still cannot produce valid code after three retries, we discard that visual idea. II. For each synthesized image, we employed Llava-1.5~\citep{liu2023improved} to check the image aesthetics, including whether visual elements within the image interfere with each other, the reasonableness of the layout, and the legibility of any text. These rules allowed us to filter out aesthetically unpleasing images. III. To ensure answer accuracy, we adopted the self-consistency~\citep{Wang2022SelfConsistencyIC} for answer generation: instructing the LLM to generate multiple responses based on the idea, code, and question, and then selecting the final answer through a voting process. 

\paragraph{Human Evaluation}
we also conduct a manual evaluation of the dataset. First, we randomly sampled 10\% of the <question, answer> pairs from our benchmarks and invited 4 graduate students in the computer science field for manual evaluation. For each sample, we designed four evaluation criteria: \textbf{Image Aesthetics, Question Rationality, Answer Accuracy, and Image-Instruction Relevance}. The criteria for Image Aesthetics and Answer Accuracy are scored from 1 to 5 (5 being the highest), while Question Rationality and Image-Instruction Relevance are divided into three levels {1, 3, 5}. The scoring criteria for each dimension and the final results of the human evaluation are shown in~\Cref{human evaluation} and~\Cref{human evaluation results}.

\begin{table}[t!]\small
\setlength\tabcolsep{0.9pt} 
\centering
\begin{tabular}{lccc}
\toprule[1pt]
\multirow{2}*{\textbf{LMMs}} & \multicolumn{3}{c}{\textbf{Acc (\%)}}\\
& \textbf{Chart} & \textbf{Table} & \textbf{Road Map} \\
\midrule

GPT-4-Vision-1106 & \textbf{50.6} & \textbf{75.8} & 23.3 \\
Claude-3-Sonnet & 46.4 & 68.4 & 38.3 \\
Qwen-VL-Plus-70B & 40.1 & 51.6 & 18.6 \\ 
\midrule
Vanilla Llava-1.5-7B & 10.5 & 15.8 & 0.3 \\
Vanilla Llava-1.5-13B & 13.4 & 18.3 & 5.1 \\
InstructBLIP-7B & 8.8 & 7.7 & 0.4 \\
InstructBLIP-13B & 2.8 & 2.1 & 0.6  \\
Deepseek-VL-Chat-1.3B & 18.4 & 24.2 & 9.6 \\
Deepseek-VL-Chat-7B & 25.2 & 31.1 & 18.8 \\
\midrule
Llava-our-62$k$ & 30.3 \tiny{$\uparrow$19.8} & 51.8 \tiny{$\uparrow$36.0} & \textbf{67.7} \tiny{$\uparrow$67.4} \\
\bottomrule
\end{tabular}
\caption{Our model is fine-tuned on chart, table, and roadmap tasks. The arrows indicate the improvements compared to Vanilla Llava-1.5-7B.}\label{sft-table1}
\end{table}

\begin{table}[t!]\small
\setlength\tabcolsep{3pt} 
\centering
\begin{tabular}{lcccc}
\toprule[1pt]
 \textbf{Data Selection} & \textbf{Size} & \textbf{Chart}  & \textbf{Table} & \textbf{Map}\\ \midrule
Vanilla Llava & 0 &  10.5 & 15.8 & 0.3 \\ \midrule
$w/$ Chart & 34.5$k$ & 29.8 & 26.7  & 8.9  \\
$w/$ Table& 10.8$k$& 17.3 & 47.8  & 6.0 \\
$w/$ Map& 17$k$ & 9.8 & 10.3 & 62.0 \\
$w/$ Chart, Table  & 45.3$k$ & 31.0  & 50.4  & 7.6 \\
$w/$ Chart, Table, Map  & 62.3$k$ & 30.3 & 51.8  & 67.7 \\
\bottomrule
\end{tabular}
\caption{We investigate the synergistic effects between the three tasks (acc in \%). Chart and table corpus can improve each other and both benefit road map tasks.}\label{three task together}
\end{table}

\begin{table*}[h!]\small
\setlength\tabcolsep{5pt} 
\centering
\begin{tabular}{lcc|ccccc}
\toprule[1pt]
\multirow{2}*{\textbf{LLM}} & \multicolumn{2}{c|}{\textbf{Weak-related Tasks (\%)}}   & \multicolumn{5}{c}{\textbf{Our Synthetic Benchmark (\%)}} \\
& ChartQA & MathVista  & Dashboard &  Relation Graph & Flowchart & Visual Puzzle & Planar Layout \\ \midrule
Vanilla Llava   & 19.9 & 25.1 & 16.5 & 29.6 & 9.6  & 3.4 & 37.7    \\
Llava-our-62$k$ & 23.9 \tiny{$\uparrow$4} & 25.9 \tiny{$\uparrow$0.8}& 16.5 & 30.1 \tiny{$\uparrow$0.5}& 12.3 \tiny{$\uparrow$2.7}& 3.6 \tiny{$\uparrow$0.2}& 44.1   \tiny{$\uparrow$6.4} \\
\bottomrule
\end{tabular}
\caption{We used two weakly related tasks and our synthetic benchmarks from five untrained tasks to evaluate the generalization capability of our $62k$ model, which was fine-tuned solely on chart, table, and road map tasks.}\label{generalization table}
\end{table*}

\section{Experiments}
First, we evaluate the performance of many leading LMMs using our benchmark containing all tasks in~\Cref{benchmark LMM}. Next, we perform instruction fine-tuning on the Llava-1.5-7B using 62,476 charts, tables, and road map instructions (denoted as Llava-our-$62k$). Then, we discuss the in-domain performance Llava-our-$62k$ and the impact of the quantity of synthetic data (\Cref{sft llava}). Lastly, we investigate whether it can be generalized to other reasoning tasks (\Cref{Generalization}).

\subsection{Settings}
We evaluated the performance of mainstream open-source and closed-source LMMs, including Llava-1.5-7B~\citep{liu2023improved}, Llava-1.5-13B, InstructBLIP-7B~\citep{dai2024instructblip}, InstructBLIP-13B, Deepseek-VL-Chat-1.3B~\citep{lu2024deepseek}, Deepseek-VL-Chat-7B, Claude-3.5-Sonnet, Claude-3-Sonnet, GPT-4o, GPT-4-Vision-1106~\citep{gpt4}, Gemini-1.5-pro, MiniCPM-2.6~\citep{yao2024minicpm}, and Qwen-VL-Plus~\citep{bai2023qwen_vl}. All models were evaluated using the same prompts and temperature settings. We provide the evaluation metrics and other training details in ~\Cref{experiments detail}.

\subsection{Benchmarking LMM's Visual Reasoning} \label{benchmark LMM}
As shown~\Cref{figure1}, we evaluate the performance of many LMMs, Llava-our-$62k$ across eight tasks, i.e., chart, table, road map, dashboard, relation graph, flowchart, visual puzzle, and planar layout. Additionally, we invited two undergraduate students to test on our benchmark. Their scores were then averaged to represent the human-level performance. The detailed results are shown in~\Cref{benchmark results}.

\paragraph{Underwhelming Abstract Image Comprehension.} We observe that for these abstract images, even advanced LMMs like GPT-4o and Claude-3.5-Sonnet achieved only 64.7\% and 59.9\% accuracy on average for all tasks, leaving a significant gap to human-level performance (82.1\%). Surprisingly, some tasks that seem straightforward for humans, such as planning a route on a map and recognizing clocks, prove challenging for LMMs. Specifically, in the dashboard task, even the best LMMs only achieved an accuracy of 54.79\% (GPT-4o). In the chart and relation graph tasks, we observe that LMMs often make errors when dealing with abstract concepts and spatial relationships. For example, in the Planar Layout task, GPT-4v often fails to distinguish the size of the three bedrooms accurately and whether they contain a washroom. These results indicate that despite significant progress in understanding semantic-rich natural photos, current LMMs still possess only a rudimentary understanding of abstract images and concepts.

\paragraph{Significant Disparity in Visual Reasoning Ability Among LMMs.} In the road map navigation task, LMMs need to dynamically plan reasonable paths based on visual input. In the visual puzzle task, LMMs should carefully observe the given diagrams, induce visual patterns, and then perform reasoning. For these two tasks, we observed a significant performance disparity between open-source and closed-source LMMs. For example, Claude-3.5-Sonnet achieved 59.2\% and 62.3\% for road map and visual puzzles, respectively, while smaller open-source models all achieved very low accuracy ($\leq$ 20\%). This disparity between open-source and closed-source LMMs is particularly pronounced in these visual reasoning tasks.

\subsection{Main Results After Fine-tuning}  \label{sft llava}
In addition to constructing the benchmark, we fine-tuned the Llava-1.5-7B model using the training sets from chart, table, and map tasks, and compared its performance with other baselines.

\paragraph{In-domain Performance.} First, as shown in \Cref{sft-table1}, compared to vanilla Llava-1.5-7B, we significantly improved its chart understanding capabilities by 19.8\% and 36\%, and also achieved the best performance in the road map navigation task (67.7\%), far surpassing closed-source LMMs like GPT-4 (23.3\%) and Claude-3 (38.3\%). Notably, we only use 68$k$ synthetic data and 4 hours of LoRA fine-tuning, elevating the chart understanding capability of Llava-1.5-7B to the Qwen-VL-Plus level. This demonstrates the tremendous potential of our synthetic data. Besides, we observe that most LMMs perform poorly on the road map navigation task, but can quickly improve after fine-tuning using our data. This highlights that current LMMs are not well-aligned in these reasoning scenarios.


\paragraph{Synergy Between Chart, Table and Road Map.}
We also studied the synergistic effects among the three tasks, such as whether chart training data benefits table and road map navigation tasks. As shown in \Cref{three task together}, we trained separately on the chart ($34.5k$), table ($10.8k$), and roadmap ($17k$) datasets. Then, we train with a mix of chart and table data, and finally with a mix of all three tasks. We found that training on charts and tables does have a positive effect on road map tasks. For example, training solely on charts or tables can lead to approximately a +5\% performance improvement in road map tasks, despite the significant differences in task types. Interestingly, the reverse is not true. The training process on road maps does not have a significant impact on chart and table tasks. We speculate that this may be due to the different capabilities required for each task.



\paragraph{Impact of Synthetic Data Quantity.} To investigate the impact of synthetic data quantity, we fine-tuned the Llava-1.5-7B model using 35$k$, 47$k$, and 62$k$ synthetic instructions respectively. As shown in~\Cref{datasize}, we observe that as the quantity of synthetic data increases, the model's performance steadily improves without reaching a plateau, especially in the math reasoning sub-task. Specifically, the accuracy for chart tasks increased from 25.78\% to 29.5\%, and the table accuracy improved by 5.4\%. These results indicate that our synthetic data are of high quality and diversity.

\subsection{Generalized to Untrained Tasks}\label{Generalization}
We evaluate whether Llava-our-62$k$ can generalize to other benchmarks, especially the tasks with significant differences. We use 1) two weakly correlated tasks: ChartQA~\citep{masry2022chartqa}, MathVista~\citep{lu2023mathvista}, and 2) our synthetic benchmarks from other five reasoning tasks. As shown in~\Cref{generalization table},  we observe that although our 62$k$ model is only trained on chart, table, and road map data, it also demonstrates improvements in other benchmarks, including chartQA (+4\%), MathVista (+0.8\%), and our synthetic benchmarks (+1.95\% on average). These results show that our model can generalize to other types of visual reasoning tasks, rather than merely fitting to the training scenarios.

\subsection{Discussion}
More than just for instruction fine-tuning, we believe abstract image comprehension capabilities can be enhanced through various strategies:

\textbf{Designing More Versatile Visual Encoders}: First, we observe that current LMMs have weak visual representation of abstract images, which may be caused by visual encoders. Most of them use clip-based encoders, which emphasize semantic features while neglecting purely visual features. It's promising to explore another visual encoder to enhance understanding of abstract images.

\textbf{Increasing Image Resolution}: Then we observed that most LMMs resize the original image to a resolution of 336x336, as it reduces the number of visual tokens. However, for these abstract images composed of lines and geometric shapes, lowering the resolution results in the loss of a significant amount of geometric features. Increasing image resolution may be a good solution.


\textbf{Investigate Relationships between Abstract Image Tasks}: Lastly, we will investigate the relationships between different abstract image tasks, quantitatively analyzing their mutual influences and their impact on LMM's abilities such as abstract image perception, spatial reasoning, and visual-symbol induction. These fine-grained studies will guide us in designing more useful abstract image tasks using our pipeline. 

\section{Conclusions}
We observe that current LMMs perform sub-optimally in perceiving and reasoning with abstract images, often failing at simple daily tasks. Therefore, we design a multimodal self-instruct strategy, enabling LLMs to autonomously synthesize various diagrams, instrument dashboards, and visual puzzles using code, and self-propose reasoning Q\&A. We synthesized $11k$ data to benchmark the current LMMs. Evaluation results underscore the significant challenges posed by our benchmark. We also synthesized $62k$ chart and road map training instructions to fine-tune a Llava-7B, enhancing its chart interpretation and map navigation abilities.

\section*{Limitations}
Our multi-modal self-instruct can synthesize a vast amount of abstract images and reasoning instructions for LLMs. However, we want to highlight that there remain some limitations or areas for improvement: 1. Our data synthesis process relies on the code generation and reasoning capabilities of LLMs, which are only available in some strong LLMs like GPT-4. Using these models is costly. It is promising to employ some advanced open-source LLMs, e.g., Qwen2, LLama3.2, to synthesize data. 2. This work used code to synthesize abstract images in eight scenarios. In the future, we can expand to more scenarios, thereby producing a massive amount of data.

\section{Acknowledgments}
We appreciate the support of \emph{Kerui Zhang} for the human evaluation of our synthetic data and website design of our project. 

This work is supported by the "Pioneer" and "Leading Goose" R\&D Programs of Zhejiang (No. 2024C03255), the National Natural Science Foundation of China (No. 62376245), the Fundamental Research Funds for the Central Universities (226-2024-00170),  the project of the Donghai Laboratory (Grant no. DH-2022ZY0013), National Key Research and Development Project of China (No. 2018AAA0101900), and MOE Engineering Research Center of Digital Library. 

\bibliography{acl_latex}

\appendix

\clearpage
\renewcommand\thefigure{\Alph{section}\arabic{figure}}    
\setcounter{figure}{0}    
\renewcommand\thetable{\Alph{section}\arabic{table}}    
\setcounter{table}{0}   

\section{Experiments Details}\label{experiments detail}
\textbf{Metrics.} Considering the diversity of output formats, including numerical values, single phrases, and long sentences, we employed different evaluation metrics. For numerical questions in chart, table, and dashboard tasks, answers within a 5\% error margin are considered correct. For numerical questions in other tasks, the predicted values must match the labeled values exactly. For single-phrase answers, the predictions should either precisely match or contain the labeled answers. For long-sentence answers, we used the Rouge-L score as the evaluation metric. For the map navigation task, we evaluated the predicted paths by calculating the Landmark Coverage Rate (LCR(\%)): we first extracted the predicted landmark sequence from the LMM's response and then compared it sequentially with the annotated landmarks sequence, calculating the proportion of correctly ordered landmarks.

\textbf{Training Details.} We fine-tuned the Llava-1.5-7B using LoRA~\citep{hu2021lora} (denoted as Llava-our-62$k$) on chart, table, and road map training sets for 1 epoch, with a batch size of 16, a learning rate of 2e-4, a rank of 128 and alpha of 256. All other parameters were kept consistent with those of Llava-1.5-7B. For reasoning questions, we concatenated the answer and rationale for instruction-following training.

\begin{table*}[!h]\small
\setlength\tabcolsep{2pt}
\centering
\begin{tabular}{lcccccccc|l}
\toprule
\multirow{2}{*}{LLMs} & \multicolumn{9}{c}{Acc (\%)} \\
 & Chart & Table & \begin{tabular}[c]{@{}c@{}}Road Map\end{tabular} & \begin{tabular}[c]{@{}c@{}}Dashboard\end{tabular} & \begin{tabular}[c]{@{}c@{}}Relation\ Graph\end{tabular} & Flowchart & \begin{tabular}[c]{@{}c@{}}Visual Puzzles\end{tabular} & \begin{tabular}[c]{@{}c@{}}Layout\end{tabular} & Avg. \\ \midrule
Human & \textbf{93.5} & \textbf{95.1} & \textbf{75.0} & \textbf{85.3} & \textbf{82.5} & \textbf{65.5} & \textbf{62.5} & \textbf{97.6} & \textbf{82.1} \\ \midrule
Claude-3.5-Sonnet & 67.24$^{\ast}$ & 84.38 & 59.24 & 54.00 & 58.52$^{\ast}$ & 49.21 & 62.38$^{\ast}$ & 82.94$^{\ast}$ & 64.74$^{\ast}$ \\
GPT-4o & 61.83 & 88.76$^{\ast}$ & 37.82 & 54.79$^{\ast}$ & 54.50 & 54.31$^{\ast}$ & 45.37 & 82.54 & 59.99 \\
Claude-3-Sonnet & 46.4 & 68.4 & 38.3 & 35.4 & 56.2 & 40.3 & 47.0 & 69.1 & 50.1 \\
GPT-4V-1106 & 50.6 & 75.8 & 23.3 & 36.2 & 52.4 & 45.3 & 35.9 & 76.6 & 49.5 \\
GPT-4o-mini & 48.7 & 77.4 & 26.7	& 46.3 & 51.1 & 42.5 & 30.8 & 75.8&	49.5 \\
Claude-3-Opus & 46.73 & 67.71 & 38.26 & 38.70 & 48.78 & 35.77 & 47.26 & 65.48 & 48.59 \\
MiniCPM-2.6 & 53.6 & 74.5 & 37.6 & 29.7 & 55.6 & 36.6 & 25.9 & 71.8  & 48.1\\
internvl-2-8b &	50.3&	73.9&	27.9&	28.9&	61.3&	41.2&	23.4&	66.6&	46.7 \\
Claude-3-Haiku & 41.83 & 57.33 & 23.17 & 35.83 & 45.99 & 23.09 & 45.94 & 58.73 & 41.49 \\
Gemini-1.5-Flash & 43.61 & 64.06 & 3.71 & 39.04 & 42.09 & 36.03 & 30.81 & 69.72 & 41.13 \\ 
glm-4v-9b	     & 47.8&  70.9	& 4.4& 	34.3& 	47.0& 	39.3	& 20.2& 	63.8& 	41.0 \\
Gemini-Pro-Vision & 43.11 & 64.92 & 3.76 & 38.87 & 41.12 & 36.09 & 29.68 & 70.12 & 40.96 \\ 
Gemini-1.5-Pro & 43.41 & 63.78 & 3.77 & 38.71 & 41.85 & 35.55 & 30.62 & 69.32 & 40.88 \\ 
Qwen-VL-Plus & 40.1 & 51.6 & 18.6 & 26.4 & 52.2 & 32.5 & 32.3 & 61.5 & 39.4 \\ 
Deepseek-VL-Chat-7B & 25.2 & 31.1 & 18.8 & 18.2 & 37.6 & 20.8 & 15.0 & 47.2 & 26.7 \\
Vanilla Llava-1.5-7B & 10.5 & 15.8 & 0.3 & 16.5 & 29.6 & 9.6 & 3.4 & 37.7 & 15.4 \\ \midrule
Llava-our-$62k$ & 30.3 & 51.8 & 67.7$^{\ast}$ & 16.5 & 30.1 & 12.3 & 3.6 & 44.1 & 32.0 \\ \midrule
\end{tabular}%
\caption{Evaluating LMMs using our synthesized benchmark containing eight reasoning tasks. Bold indicates the best performance. $^{\ast}$ indicates the second highest.}
\label{benchmark results}
\end{table*}

\begin{figure}[!htb] 
\centering 
\includegraphics[width=0.5\textwidth]{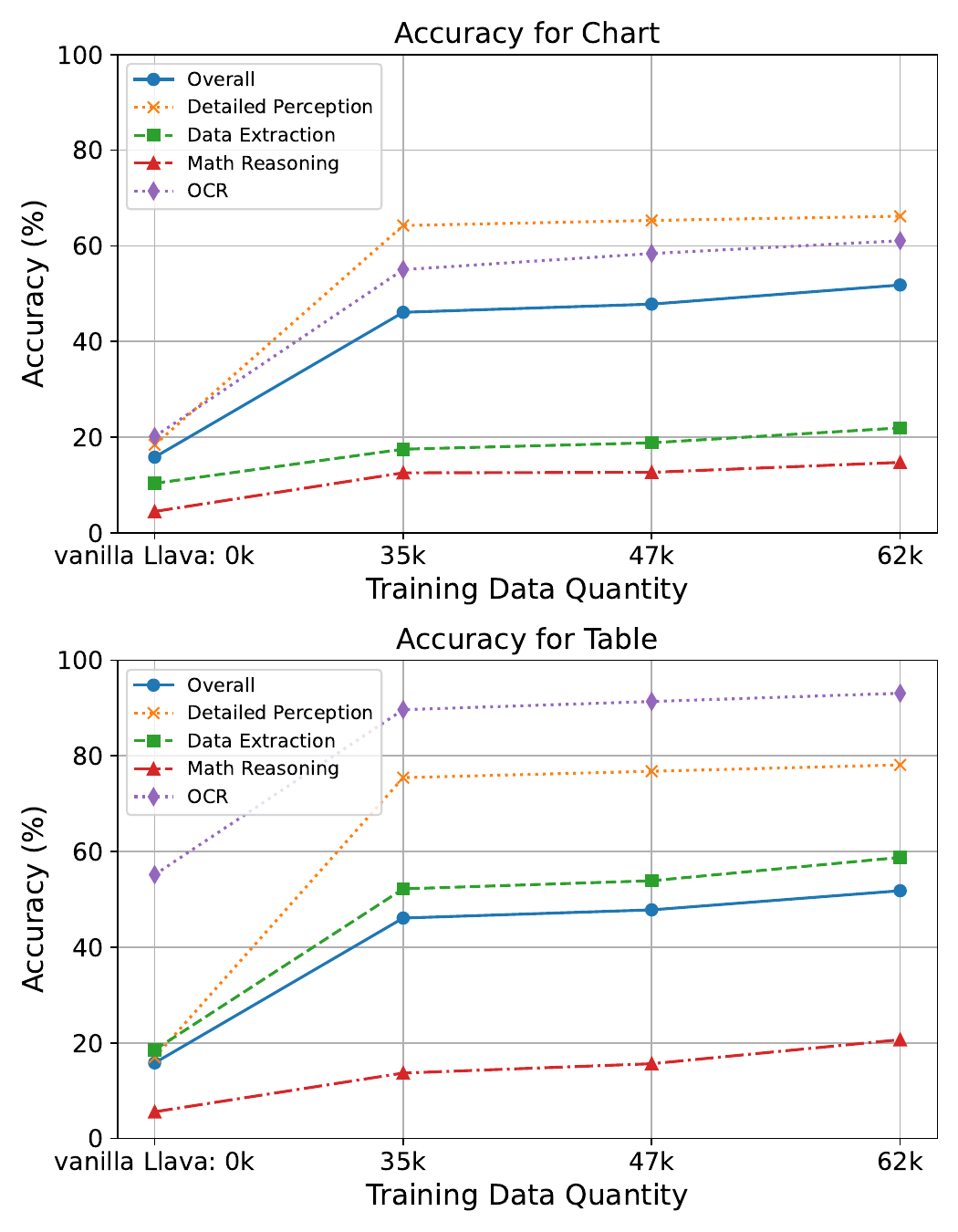} %
\caption{ We analyzed the impact of synthetic data quantity on the model's performance. We fine-tune Llava-1.5-7B using chart and table instruction data of varying scales and report its accuracy. Additionally, we report the accuracy for four sub-category tasks: Detailed Perception, Data Extraction, Math Reasoning, and OCR.}\label{datasize} 
\end{figure}

\begin{figure*}[!htb] 
\centering 
\includegraphics[width=1\textwidth]{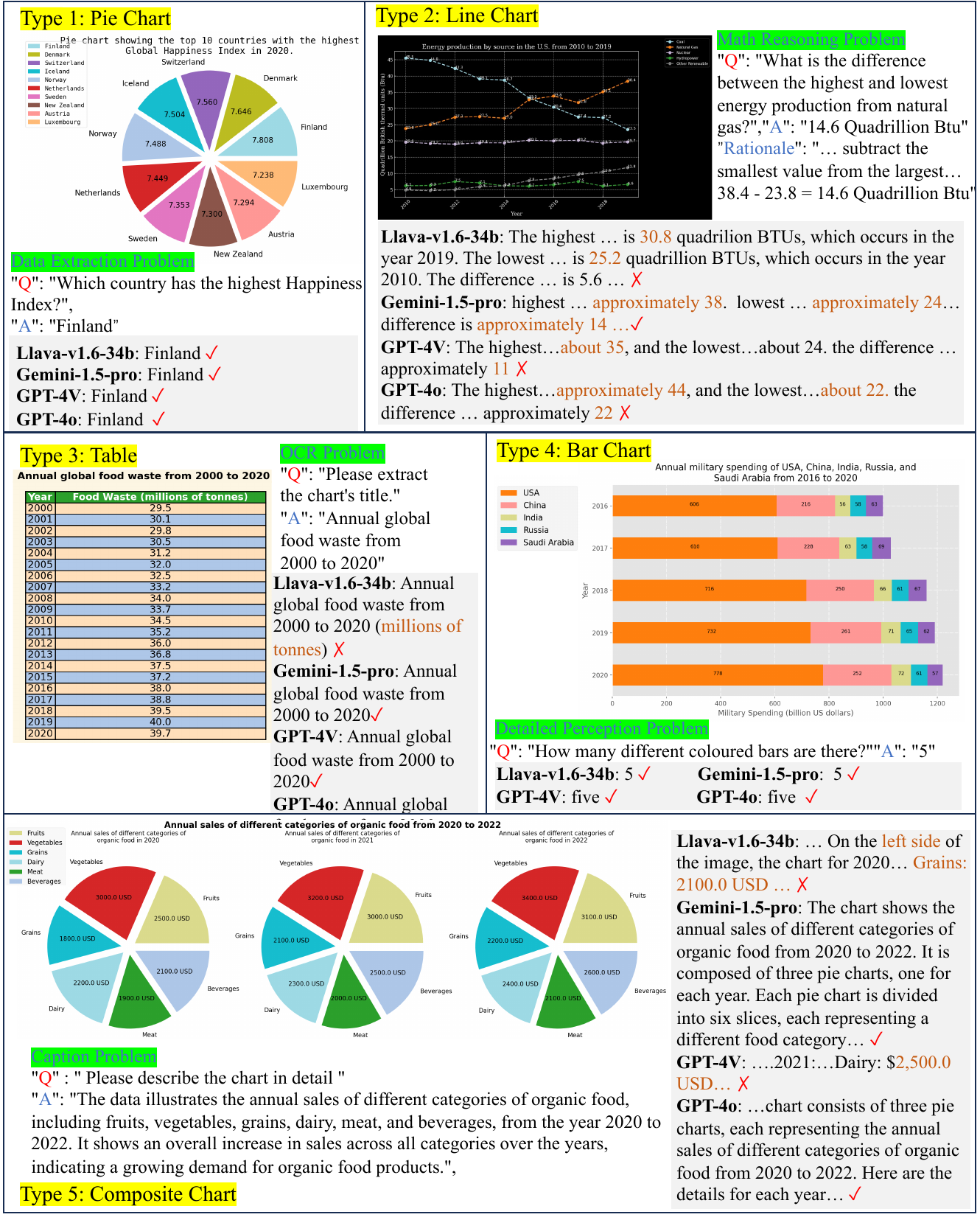} %
\caption{The chart task includes five types of charts (\textcolor{yellow}{pie chart, line chart, table, bar chart, composite chart}), each containing five types of questions (\textcolor{green}{Data Extraction, Math Reasoning, OCR, Detailed Perception, Caption Problem}).}\label{chart example} 
\end{figure*}

\begin{figure*}[!htb] 
\centering 
\includegraphics[width=1\textwidth]{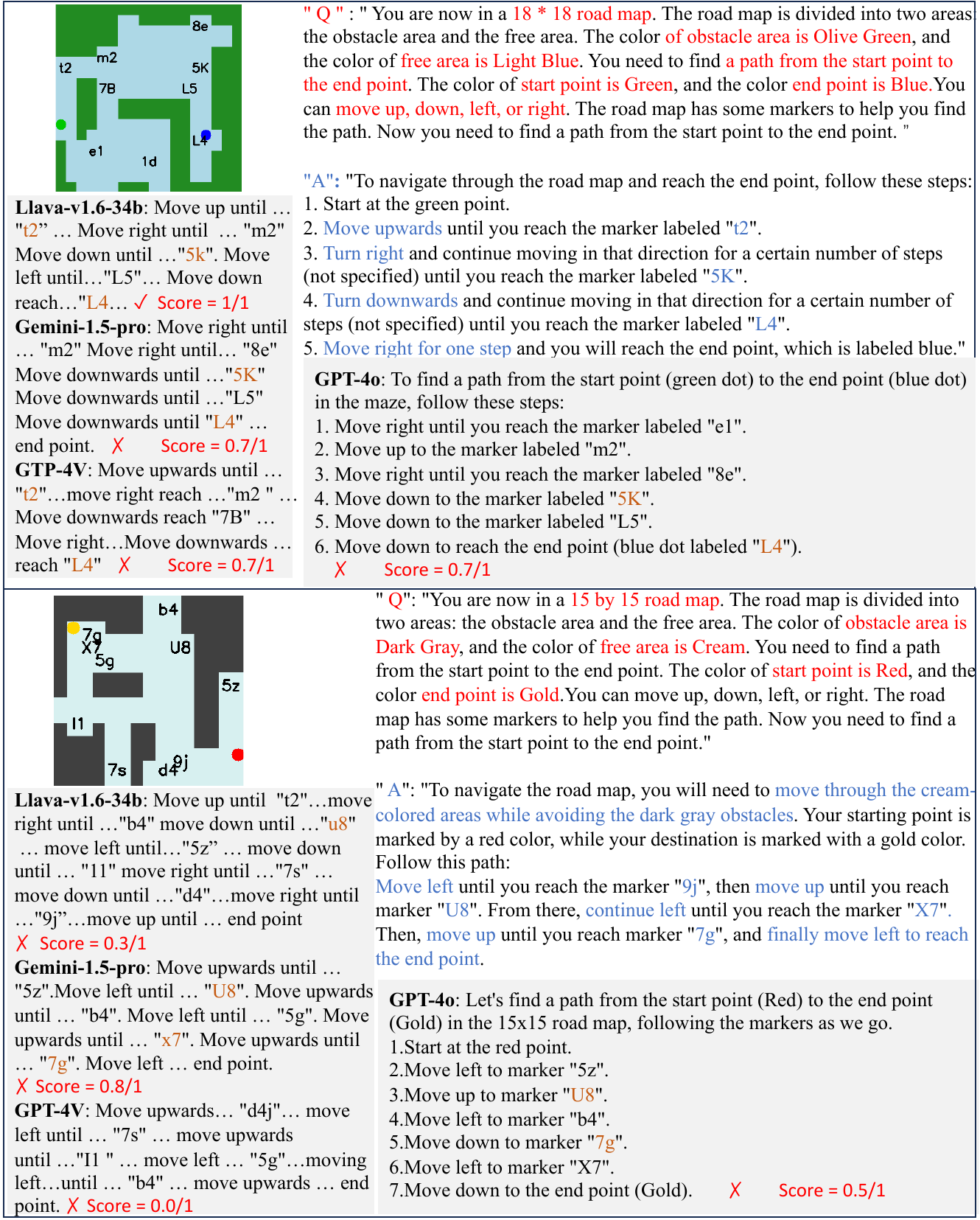} %
\caption{We present two examples of road map navigation, including the synthesized simulated maps, questions, and answers.}\label{map navigation example} 
\end{figure*}

\begin{figure*}[!htb] 
\centering 
\includegraphics[width=1\textwidth]{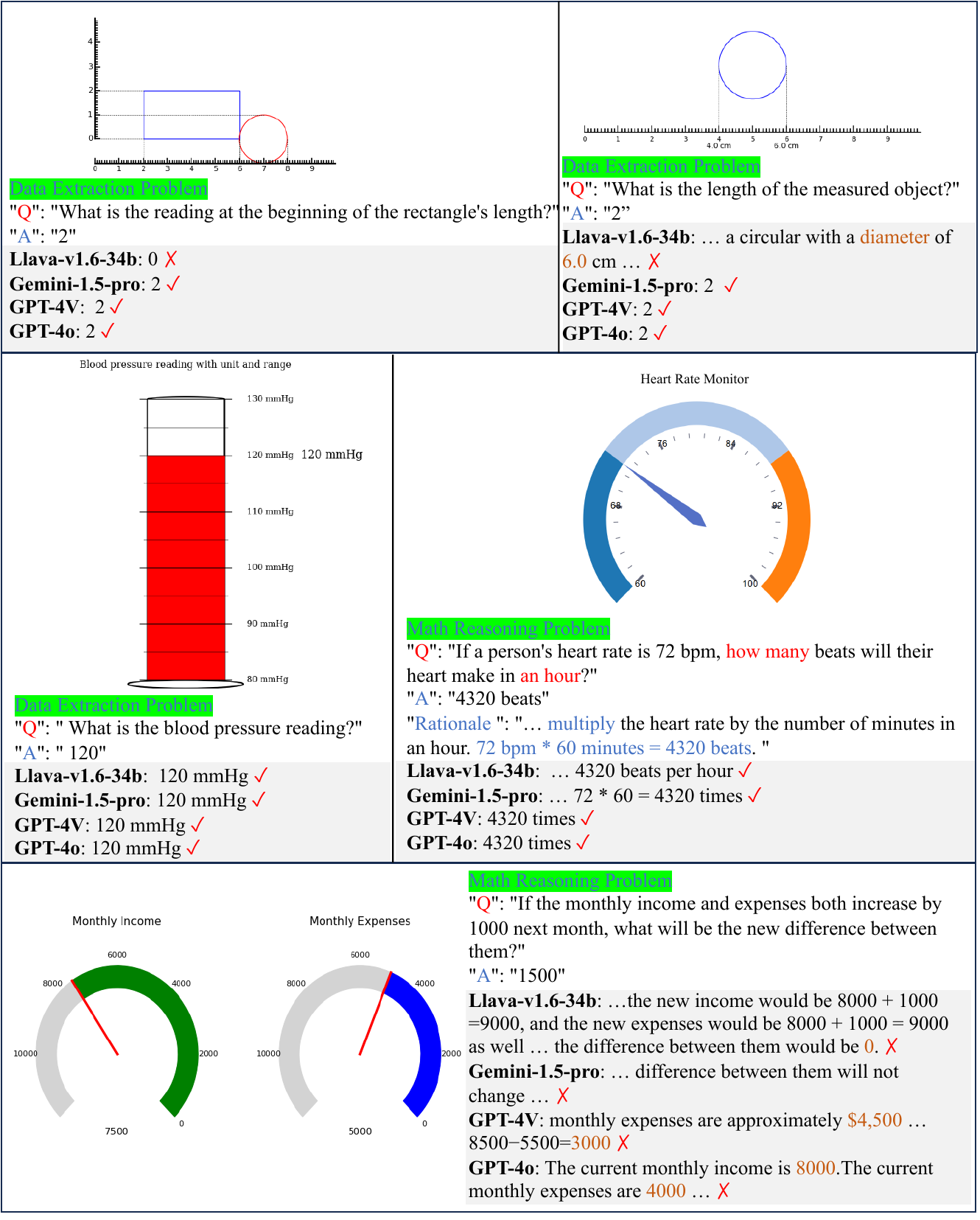} %
\caption{We present five examples of dashboard.}\label{map navigation example} 
\end{figure*}

\begin{figure*}[!htb] 
\centering 
\includegraphics[width=1\textwidth]{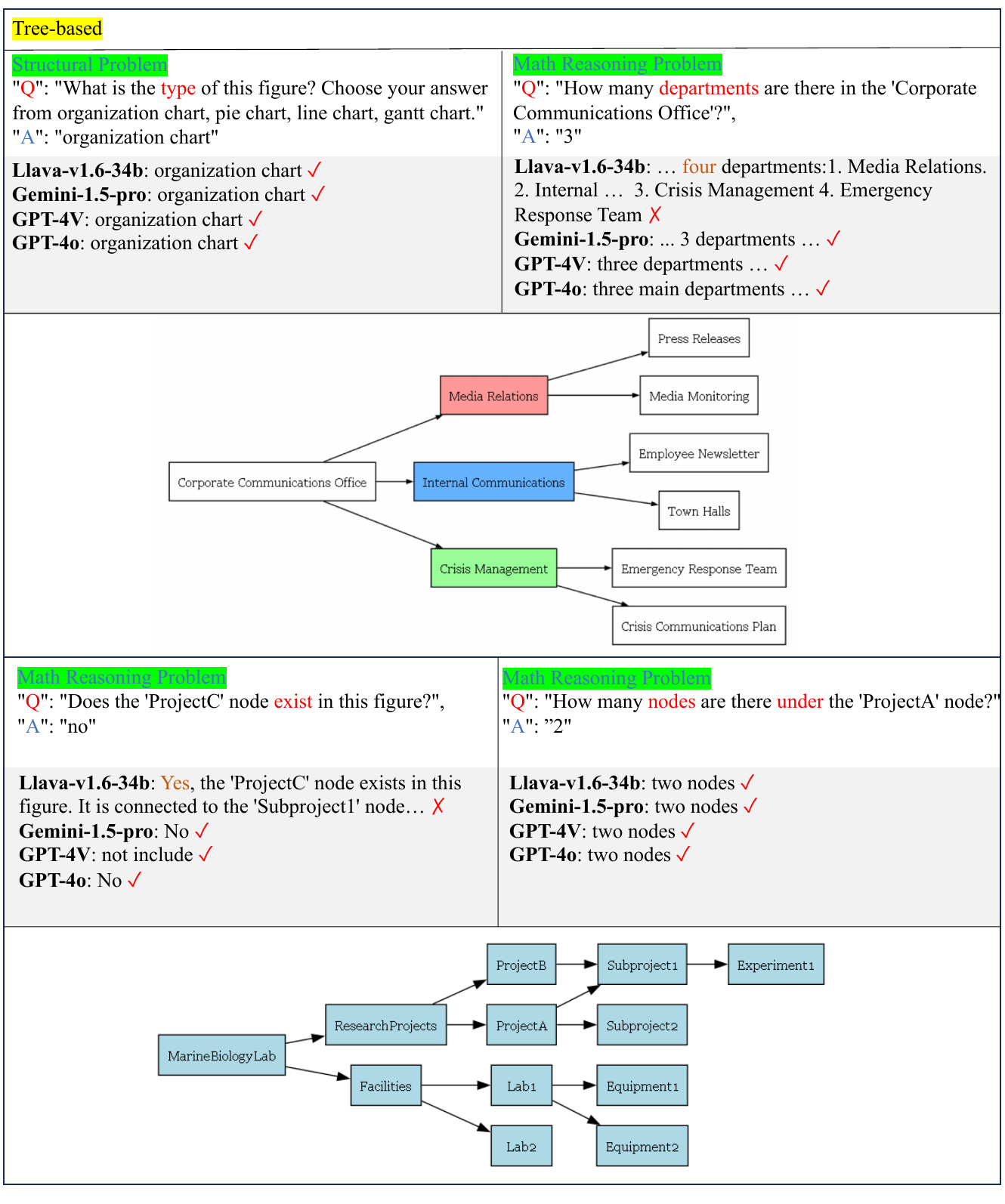} %
\caption{We present two examples of relation graph, each containing two types of questions.}\label{relation graph example} 
\end{figure*}

\begin{figure*}[!htb] 
\centering 
\includegraphics[width=1\textwidth]{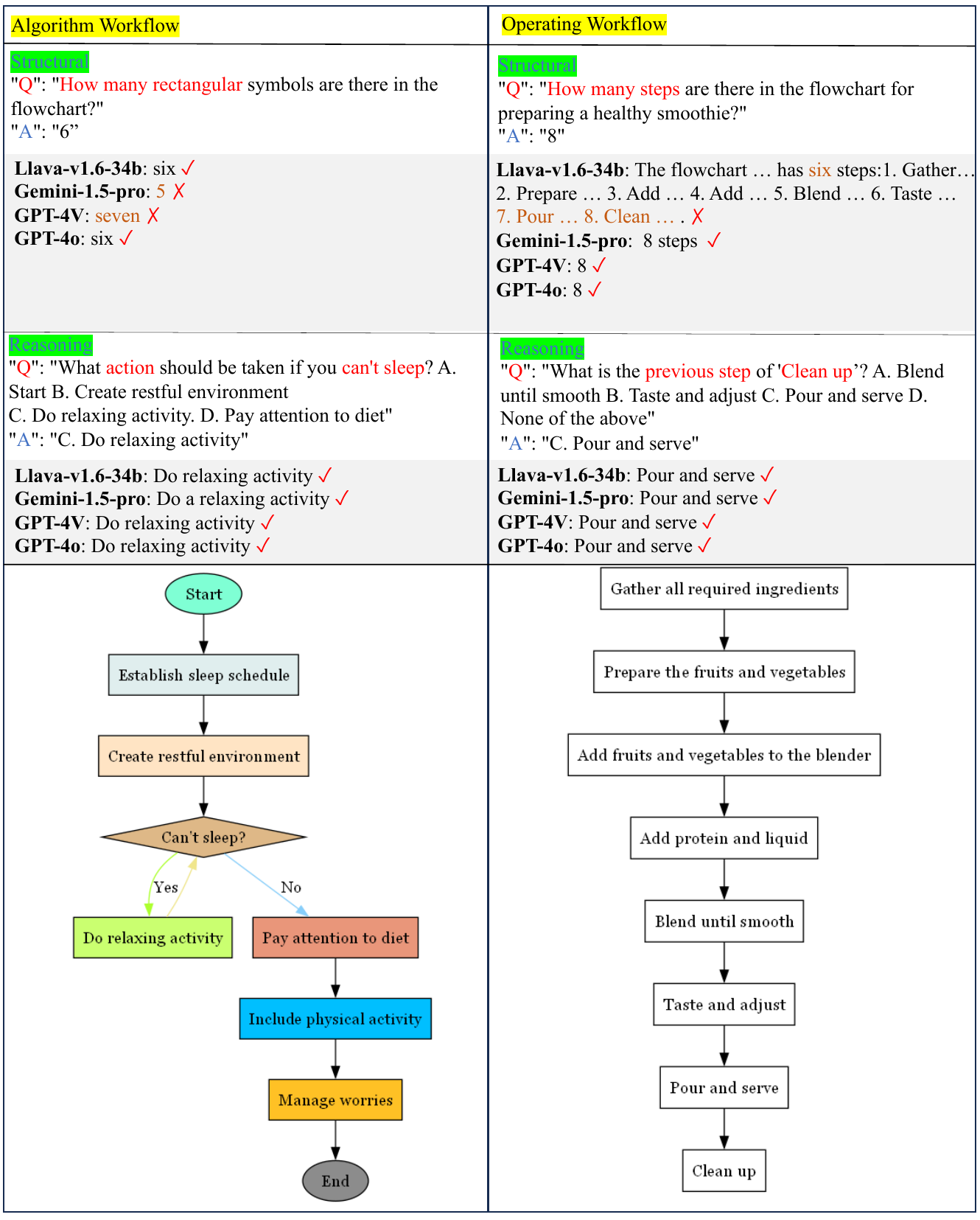} %
\caption{We present two examples of flowchart (\textcolor{yellow}{algorithm workflow and operating workflow}), each containing two kinds of questions (\textcolor{green}{Structural and Reasoning Problem}).}\label{flowchart example} 
\end{figure*}

\begin{figure*}[!htb] 
\centering 
\includegraphics[width=1\textwidth]{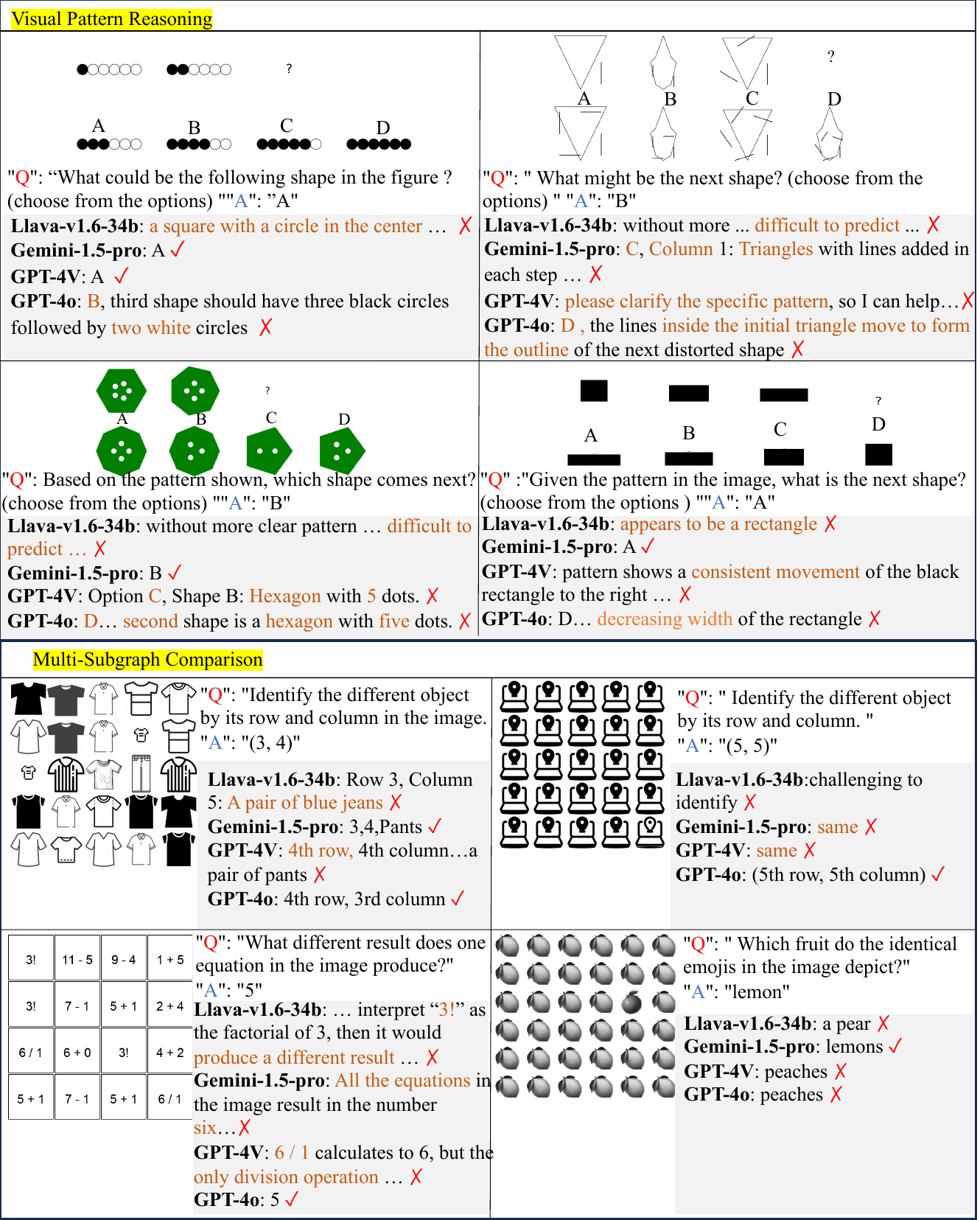} %
\caption{We present two categories of visual puzzles (\textcolor{yellow}{visual pattern reasoning and muti-subgraph comparison}), each containing four visual puzzle graphs, questions, and answers.}\label{visual puzzle example} 
\end{figure*}

\begin{figure*}[!htb] 
\centering 
\includegraphics[width=1\textwidth]{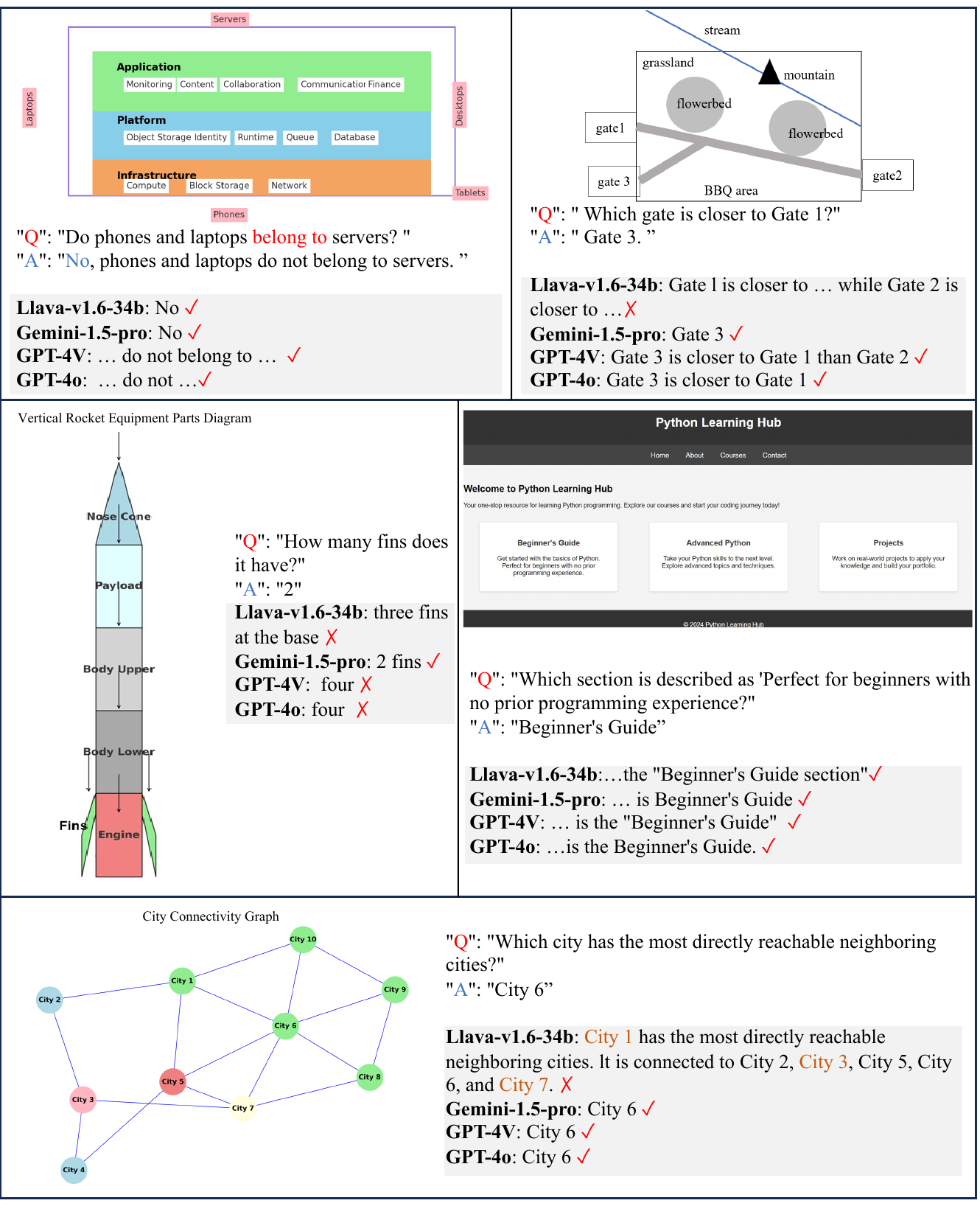} %
\caption{We present five examples of 2D planar layout, including the layout graph, problems, answers and rationales.}\label{layout example} 
\end{figure*}

\begin{figure*}\label{detail prompt}
\begin{lstlisting}

--------------- (*@\textbf{Data Prompt}@*) ---------------
Generate data related to (*@\color{blue}{Digital Forensics Unit}@*).
Requirements:
The data should describe a tree-like structure of Digital Forensics Unit.
There can be multiple layers and certain nodes can have no children.
The data should not contain too much nodes and should not be too complicated.
Increase the depth of the data, but no more than 3 nodes in the same layer.
The total number of nodes should not exceed 8.
Output format: {"data": {...}}

(*@\color{green}{Instance:}@*)
{
  "data": {
    "Digital Forensics Unit": {
    "Case Management": {
      "Evidence Collection": {},
      "Analysis": {}
    },
    "Training and Development": {
      "Workshops": {},
      "Certifications": {}
    }
  }
}

--------------- (*@\textbf{Title Prompt}@*) ---------------
Generate a title for the data.
Requirements:
The title should be brief and concise.
The title should describe the general content of the data.
Output format: {"caption": "..." }

(*@\color{green}{Instance:}@*) Digital Forensics Unit

--------------- (*@\textbf{Code Prompt}@*) ---------------
Generate high quality (*@\color{blue}{python code}@*) to draw a organization chart for the data.
Requirements:
The code should only use packages from ['graphviz'].
The code must conform general requirements (given in JSON format):
{
  "title": "Graphic Design Team",
  "data": [
    "all data must be used",
    "annotate the node on the organization chart"
  ],
  "layout": [
    "draw an hierarchy structured organization chart of the data",
    "nodes different levels are positioned vertically, nodes on the same level are positioned horizontallyuse arrows or lines to connect nodes",
    "do not show axis"
  ]
}
Output format: ```python ... ```

\end{lstlisting}
\centering
\includegraphics[width=0.8\textwidth]{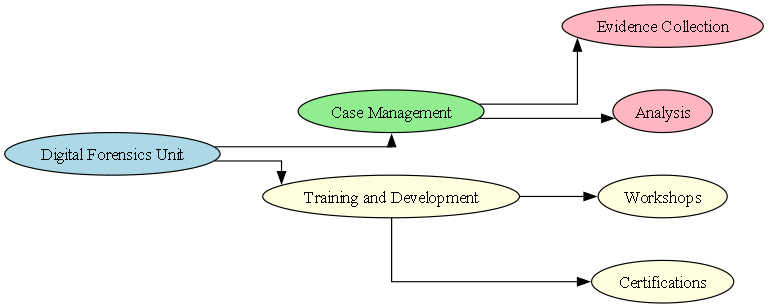}
\end{figure*}

\begin{figure*}
\begin{lstlisting}
(continue from last page)

--------------- (*@\textbf{Question-Answer Prompt}@*) ---------------
Generate correct and high quality (*@\color{blue}{question-answer}@*) pairs about the data and the organization chart.
Requirements:
Question-answer types: 
{
  (*@\color{blue}{STRUCTURAL}@*): {
    'Example 1': 'What is the type of this figure? Choose your answer from organization chart, pie chart, line chart, gantt chart.', 
    'Example 2': "What's the color of {node}?"}, 
  (*@\color{blue}{MATH\_REASONING}@*): {
    'Example 1': 'Does {name} node exist in this figure?', 
    'Example 2': 'How many nodes are there?'}
}
If applicable, the answer can be a single word.
Consider the data and code together to get the answer.
Output format: {
    "STRUCTURAL":[{"Q":"...", "A":"..."}, ...],
    "MATH_REASONING":[{"Q":"...", "A":"..."}, ...]
}

(*@\color{green}{Instance:}@*)
{
      "STRUCTURAL": [
        {
          "Q": "What is the type of this figure? Choose your answer from organization chart, pie chart, line chart, gantt chart.",
          "A": "organization chart"
        },
        {
          "Q": "What's the color of the 'Digital Forensics Unit' node?",
          "A": "lightblue"
        }
      ],
      "MATH_REASONING": [
        {
          "Q": "How many nodes are there in the 'Digital Forensics Unit'?",
          "A": "2"
        },
        {
          "Q": "Does the 'Evidence Collection' node exist in this figure?",
          "A": "Yes"
        },
        {
          "Q": "How many nodes are there in the 'Case Management' department?",
          "A": "2"
        },
        {
          "Q": "How many nodes are there in the 'Training and Development' department?",
          "A": "2"
        },
        {
          "Q": "How many departments are there in the 'Digital Forensics Unit'?",
          "A": "2"
        }
    }
}

\end{lstlisting}
\end{figure*}
\section{Human Evaluation} \label{human evaluation}
As discussed in the paper, we design four evaluation metrics to manually assess the quality of the benchmark: Image Aesthetics, Question Rationality, Answer Accuracy, and Image–Instruction Relevance. The specific criteria are as follows:

\begin{itemize}
    \item Image Aesthetics: Are the colors appropriate, are the details clearly visible, is the spatial layout reasonable, and are there any obstructions between objects?	
    \item Question Rationality: Is the question reasonable? Is the wording clear and unambiguous? Is the difficulty level moderate, neither too simple nor too difficult?	
    \item Answer Accuracy: Is the answer accurate? Is the rationale logical? Are the answer and rationale consistent with each other?		
    \item Image–Instruction Relevance: Is the answer related to the image? Can the question be answered without looking at the image?	
\end{itemize}

We evaluated the benchmark both before and after filtering, with the results presented in \Cref{human evaluation results}. These results indicate that the quality of our benchmark has significantly improved post-filtering, particularly in terms of Image Aesthetics and Answer Accuracy.

\begin{table*}[!h]\small
\setlength\tabcolsep{6pt}
\centering
\begin{tabular}{lcccc}
\toprule
&Image Aesthetics& Question Rationality&Answer Accuracy&Image-Instruction Relevance\\ \midrule
Before Filtering &   2.4 & 3.9 & 3.5&4.5\\
After Filtering &  4.0 & 4.1 &4.3 &4.4 \\ \bottomrule
\end{tabular}%
\caption{The results of the human evaluation.}
\label{human evaluation results}
\end{table*}

\section{Additional Experiment Results}\label{additional results}
As discussed in~\Cref{benchmark LMM}, we evaluate the performance of many LMMs, Llava-our-62k 
 and humans using our benchmark. All results are shown in~\Cref{benchmark results}. Besides, as shown in~\Cref{caption subtask}, we also calculated the Rough-L score for the caption sub-task in the chart and table. 

\begin{table}[!h]\small
\setlength\tabcolsep{2pt}
\centering
\begin{tabular}{lcc}
\toprule
\multirow{2}*{\textbf{LLMs}} & \multicolumn{2}{c}{\textbf{Rough-L}}  \\ 
 & \textbf{Chart} & \textbf{Table}  \\ \midrule
GPT-4Vision-1106 & 0.42 & 0.42 \\
Claude-3-Sonnet &   0.48 & 0.46\\
Qwen-VL-Plus & 0.36 & 0.37 \\
Vanilla Llava-1.5-7B & 0.33 & 0.37\\
Vanilla Llava-1.5-13B & 0.33  & 0.40 \\
InstructBLIP-7B & 0.04 & 0.23 \\
InstructBLIP-13B & 0.05 & 0.11 \\
Deepseek-VL-Chat-1.3B & 0.36 & 0.35 \\
Deepseek-VL-Chat-7B & 0.39 & 0.37 \\
Llava-our-62k &  0.46 & 0.44 \\ \bottomrule
\end{tabular}%
\caption{For the chart and table tasks, we also calculated the captioning results.}
\label{caption subtask}
\end{table}

\section{Related Work}
\subsection{Multi-modal LLMs}
With the rapid development of Large Language Models (LLM), many researchers are currently devoting their efforts to developing multimodal large models (MLLM) for visual understanding and reasoning tasks. Beyond OpenAI's GPT-4V and Google's Gemini, numerous open-sourced MLLMs have also emerged and gained significant progress.

Recently, MLLMs commonly align visual perception with LLMs to acquire multimodal perceptions through lightweight vision-to-language adapters, including projection, Q-former and additional cross-attention layers. For example, Kosmos-1/2~\citep{Huang2023LanguageIN,Peng2023Kosmos2GM} and LLaVA-series models~\citep{liu2024visual,liu2023improved} adopt a linear layer or an MLP to project visual inputs into textual embeddings. Furthermore, PaLM-E~\citep{Driess2023PaLMEAE}, PandaGPT~\citep{Su2023PandaGPTOM}, NExT-GPT~\citep{Wu2023NExTGPTAM} and AnyGPT~\citep{Zhan2024AnyGPTUM} even project other multimodal data such as audio, video and robot sensor data into the textual embeddings. 
Q-former was first proposed in BLIP-2~\citep{li2023blip} by employing a set of learnable queries to bridge the gap between a frozen image encoder and the LLM. It has been used in several other approaches, such as LL3DA~\citep{Chen2023LL3DAVI}, minigpt-4~\citep{Zhu2023MiniGPT4EV}, InstructBLIP~\citep{dai2024instructblip} and mPLUG-Owl~\citep{Ye2023mPLUGOwlME}. Additionally, Flamingo~\citep{alayrac2022flamingo} and Otter~\citep{Li2023OtterAM} inserted additional cross-attention layers into the frozen LLM to bridge the vision-only and language-only models. 

However, those models are primarily focused on natural images, and there still remain challenges in the comprehension of complex fine-grained images such as charts, documents, and diagrams. 

\subsection{Benchmark For Multimodal Model}
Designing a fair benchmark to evaluate the capabilities of multimodal models has garnered widespread attention within the academic community\citep{antol2015vqa,Fu2023MMEAC,xu2023lvlm,liu2023mmc,yu2023mm,yue2024mmmu,liu2024seeing,tong2024eyes,huang2024olympicarena}. Recently, some multimodal benchmarks have made valuable explorations into the visual reasoning capabilities and fine-grained recognition abilities of LMMs~\citep{yin2024lamm,liu2023mmbench,ying2024mmt,li2024seed,wang2024picture,chen2024mindbench,wu2024vsp,singh2024flowvqa,zhang2024vcr}.

Besides, several MLLMs have been proposed for chart comprehension and reasoning, including ChartLlama~\citep{han2023chartllama}, Unichart~\citep{Masry2023UniChartAU}, Structchart~\citep{Xia2023StructChartPS}, FinVis-GPT~\citep{Wang2023FinVisGPTAM}, TinyChart~\citep{Zhang2024TinyChartEC}, CharXiv~\citep{wang2024charxiv}, ChartX~\citep{xia2024chartx}, TableVQA-Bench~\citep{kim2024tablevqa} and mChartQA~\citep{wei2024mchartqa}. mPLUG-DocOwl~\citep{Ye2023mPLUGDocOwlMM} strengthens the OCR-free document understanding ability with a document instruction tuning dataset. Chartassisstant~\citep{meng2024chartassisstant} undergoes a two-stage training process, starting with pre-training on chart-to-table parsing to align chart and text, followed by multitask instruction-following fine-tuning. ChartInstruct~\citep{masry2024chartinstruct} employs a
two-step approach to extract chart data tables
and input them into the LLM. These efforts have all contributed to the advancement of multimodal technologies.

\subsection{Data Synthesis}
Data synthesis is widely used in LLM training to supplement the insufficiency of instruction-following data. Many studies focus on generating high-quality synthetic data either distilling dialogue data from a strong LLM~\citep{wang2022selfinstruct,zhang-etal-2022-multi-view,zhang-etal-2023-expression,xu2023wizardlm,yu2023metamath,chen2023sharegpt4v,zhao2023genixer}, or using external tools to refine LLM-generated synthetic data~\citep{wei2023magicoder,lee2024llm2llm}. For instance, \citet{wang2022selfinstruct} proposed \emph{Self-Instruct} to improve the instruction-following ability of LLMs via their own generation of instruction data. \citet{xu2023wizardlm} further generated  more complex instruction through \emph{Evol-Instruct}.
\citet{yu2023metamath} synthesized a mathematical dataset from LLMs by bootstrapping mathematical questions and rewriting the question from multiple perspectives. \citet{wei2023magicoder} can generate diverse and realistic coding problems from open-source code snippets. \citet{Lei2024AutoCoderEC} can also create high-quality large code datasets for LLMs. It simulates programmers writing code and conducting unit tests through agent interactions, ensuring annotation accuracy with an external code executor.

\end{document}